%% file: main.tex
\documentclass[10pt,twocolumn,letterpaper]{article}

\usepackage[pagenumbers]{cvpr}
\input{preamble}

\definecolor{cvprblue}{rgb}{0.21,0.49,0.74}
\usepackage[breaklinks,colorlinks,allcolors=cvprblue]{hyperref}

\AtBeginDocument{%
  \crefname{figure}{Fig.}{Figs.}%
  \Crefname{figure}{Fig.}{Figs.}%
  \crefname{table}{Tab.}{Tabs.}%
  \Crefname{table}{Tab.}{Tabs.}%
  \crefname{section}{Sec.}{Secs.}%
  \Crefname{section}{Sec.}{Secs.}%
}

\def\confName{CVPR}
\def\confYear{2026}

\title{GS-Voxel: Fitting-Free Structured Latents for Large-Scale 3DGS Generation}

\author{
Ming Qian$^{1}$ \quad
Zijian Wang$^{1}$ \quad
Minchao Sun$^{1}$ \quad
Jincheng Xiong$^{1,2}$\\
Hang Zhang$^{1}$ \quad
Mu Xu$^{1}$ \quad
Chi Wang$^{2}$ \quad
Baoquan Chen$^{3}$\\[2pt]
$^{1}$Amap, Alibaba \qquad
$^{2}$Zhejiang University \qquad
$^{3}$Peking University
}

\makeatletter
\apptocmd{\@maketitle}{%
  \vspace*{-8pt}%
  \begin{minipage}{\textwidth}
    \input{body/fig_tex/teaser}
    \vspace*{-10pt}%
    \begin{center}
      {\large\bfseries Abstract}
    \end{center}
    \vspace*{-6pt}%
    \begin{multicols}{2}
      \itshape
      \input{body/0.abstract}
    \end{multicols}
    \footnotetext{Correspondence: \href{mailto:qianming.chain@gmail.com}{\texttt{qianming.chain@gmail.com}}}
  \end{minipage}
}{}{}
\makeatother

\begin{document}
\maketitle
\clearpage
\input{body/1.introduction}
\input{body/2.related_work}
\input{body/3.method}
\input{body/4.results}
\input{body/5.conclusion}
\input{body/6.limitations}

{
  \small
  \bibliographystyle{ieeenat_fullname}
  \bibliography{reference}
}

\clearpage
\appendix
\input{body/6.appendix}
\input{body/5_1_imagesupp}

\end{document}

%% file: preamble.tex
\usepackage{amsmath}
\usepackage{amssymb}
\usepackage{booktabs}
\usepackage{colortbl}
\usepackage{enumitem}
\usepackage{fancyvrb}
\usepackage{fvextra}
\usepackage{graphicx}
\usepackage{makecell}
\usepackage{microtype}
\usepackage{multicol}
\usepackage{pifont}
\usepackage{siunitx}
\usepackage[most]{tcolorbox}
\usepackage{xspace}

\newcommand{\ourmethod}{\textit{GS-Voxel}\xspace}

\providecommand{\Description}[1]{}

%% file: body/fig_tex/teaser.tex
\begin{center}
  \centering
  \includegraphics[width=0.92\textwidth]{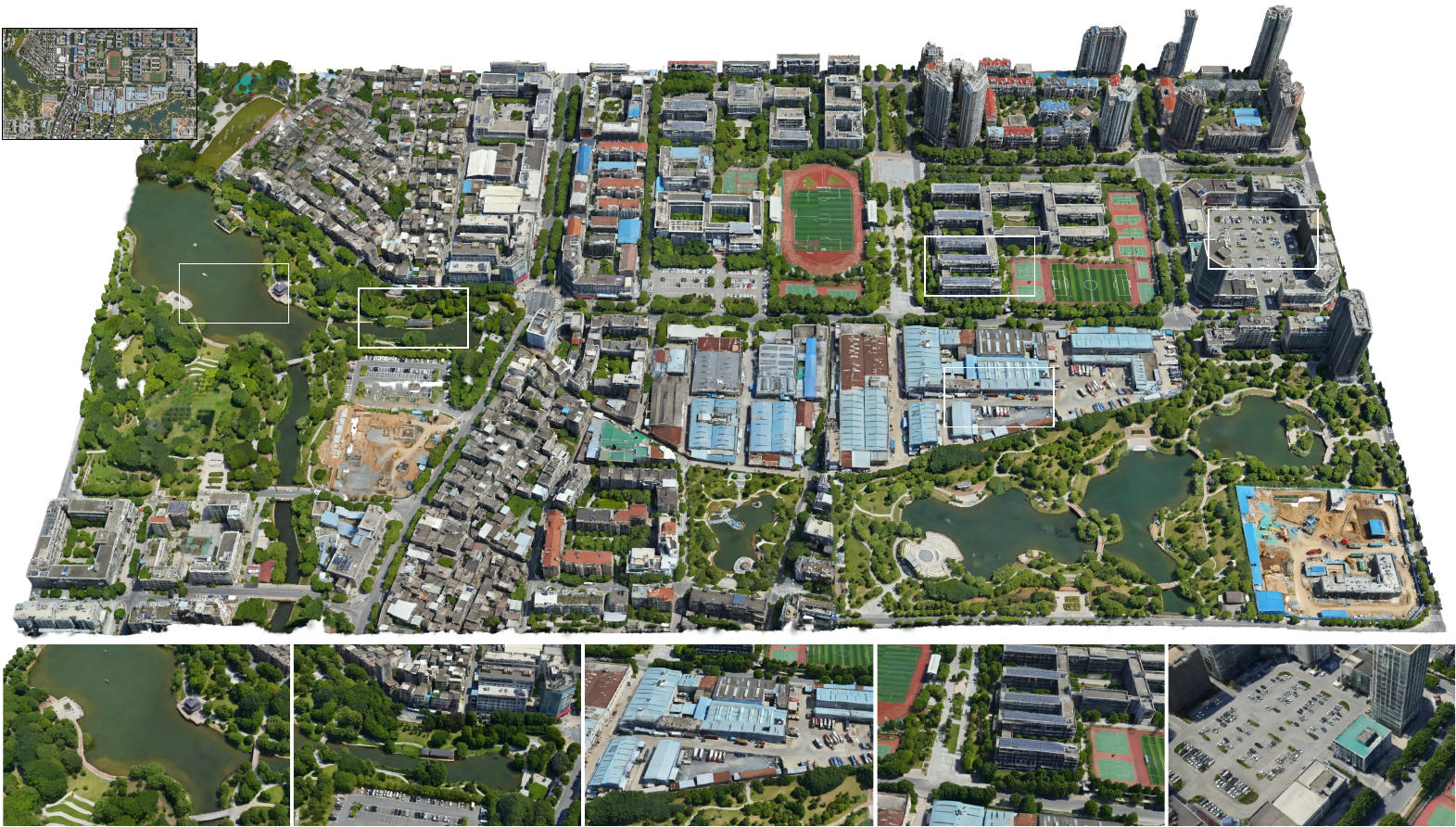}
  \Description{}
    \captionof{figure}{Given a large-footprint satellite-view conditioning image ($1{,}400\,\mathrm{m}\times800\,\mathrm{m}$, top-left), our GS-Voxel-based pipeline generates a large-area 3DGS scene. The five crops highlight local appearance and geometric detail.}
\label{teaser} 
\end{center}

%% file: body/0.abstract.tex
Many scalable latent 3D generators operate on structured tensors, whereas pre-optimized 3D Gaussian Splatting (3DGS) reconstructions are unordered, spatially irregular, and vary widely in primitive count. We present \ourmethod, a fitting-free structured latent framework, and evaluate it for large-scale aerial 3D Gaussian scene generation. GS-Voxel deterministically converts a compatible pre-optimized 3DGS reconstruction into sparse active voxels without additional per-scene optimization, retaining the sub-voxel positions and rendering attributes of the selected primitives. A GS-specific factorized VAE then separately encodes voxel geometry and local Gaussian attributes into sparse 3D latents whose size grows with the number of occupied voxels rather than being limited by a fixed scene-wide primitive count. We train image-conditioned flow models in the GS-Voxel latent space to generate aerial 3DGS scenes. A key application enabled by GS-Voxel is large-area scene generation: overlap-aware tiled inference extends synthesis beyond a single training crop conditioned on satellite-view images. Our results show that GS-Voxel provides structured latents for pre-optimized aerial 3DGS reconstructions, with latent capacity that grows with the number of occupied voxels.

%% file: body/1.introduction.tex
\section{Introduction}

\input{body/table_tex/intro_compare}

Latent 3D generative models have advanced rapidly at the object level by separating representation learning from generative modeling~\cite{xiang2025structured,xiang2025nativecompactstructuredlatents,zhang2024clay,lai2025hunyuan3d25highfidelity3d}. A compact structured latent can be modeled efficiently with diffusion or flow-based architectures and decoded into detailed 3D content. Extending this paradigm to real-world outdoor scenes, however, requires a representation that simultaneously preserves photorealistic appearance, accommodates large spatial extents, and remains tractable for generative learning.

We study this representation problem using large-scale aerial 3D Gaussian Splatting (3DGS)~\cite{3DGS} as the target domain. Such reconstructions capture detailed appearance and non-mesh-friendly content such as vegetation without requiring explicit topology. Yet a pre-optimized 3DGS reconstruction is an unordered, spatially irregular set whose cardinality varies across scenes. Raw $200\,\mathrm{m}\times200\,\mathrm{m}$ aerial 3DGS tiles can contain more than 3.0 million primitives, far beyond the global primitive budgets common in object-centric generation. Native 3DGS therefore does not directly provide the structured spatial tensor expected by the sparse latent architectures used in this work.

Existing Gaussian generation methods address irregularity in different ways. GaussianCube~\cite{zhang2024gaussiancube} refits a fixed number of Gaussians and globally rearranges them with Optimal Transport, while L3DG~\cite{roessle2024l3dg} learns latent diffusion primarily for object- and room-scale Gaussian scenes. Can3Tok~\cite{gao2025can3tok} tokenizes a globally fixed-cardinality scene-level 3DGS set into a fixed-size latent, whereas TripoSplat~\cite{yan2026triposplat} learns an adaptive density distribution and demonstrates variable-budget object decoding from 33K to 262K Gaussians. Together, these works cover globally rearranged Gaussian sets, fixed-cardinality scene latents, and adaptive object-level outputs. However, they do not directly convert an already optimized, million-scale 3DGS reconstruction into a spatial latent without global refitting or primitive rearrangement.

To address this representation bottleneck, we present \ourmethod, a fitting-free structured latent framework designed and evaluated for large-scale aerial 3D Gaussian scene generation. Here, \emph{fitting-free} refers specifically to per-scene conversion: given a pre-optimized 3DGS reconstruction in the supported SH0 attribute format, constructing GS-Voxel and encoding its latent require no additional optimization of that scene, no fitting to a fixed-cardinality scene-wide Gaussian template, and no global Optimal Transport rearrangement of the input primitives. The conversion is deterministic, while the factorized VAE and image-conditioned flow models are trained once across the dataset. GS-Voxel voxelizes the input set, retains the selected primitives' sub-voxel positions and rendering attributes, and imposes a fixed slot budget only within each active voxel. The factorized VAE separately encodes voxel geometry and local Gaussian attributes into structured sparse 3D latents. We evaluate GS-Voxel by training image-conditioned flow models in its latent space to generate aerial 3DGS scenes. We further demonstrate large-area synthesis through overlap-aware tiled inference.

%% file: body/table_tex/intro_compare.tex
\begin{table*}[t]
\centering
\caption{Taxonomic comparison of representative outdoor 3D scene creation systems.}
\label{tab:comparison}
\vspace{-0.5em}
\resizebox{\textwidth}{!}{%
\begin{tabular}{l l l l l l}
\toprule
\textbf{Method} & \textbf{Training Data} & \textbf{Condition} & \textbf{Beyond Buildings} & \textbf{Large-Scale} & \textbf{Core Paradigm} \\
\midrule
Sat2Scene~\cite{sat2scene} & Images \& Height Map & Layout & \textcolor{red}{\ding{55}} & \textcolor{red}{\ding{55}} & Geo. Coloring \\
Sat2City~\cite{sat2city} & Mesh & Height Map & \textcolor{red}{\ding{55}} & {\color{green}\checkmark} & Building Object Gen. \\
Sat2Density++~\cite{Qian_2026_Sat2Densitypp} & Images & Single Satellite & {\color{green}\checkmark} & \textcolor{red}{\ding{55}} & Feedforward \\
Sat3DGen~\cite{qian2026satdgen} & Images & Single Satellite & {\color{green}\checkmark} & \textcolor{red}{\ding{55}} & Feedforward \\
UrbanWorld~\cite{urbanworld} & Mesh + UV & OSM or Layout & {\color{green}\checkmark} & {\color{green}\checkmark} & Multi-Stage \\
SynCity~\cite{engstler2025syncity} & N/A (training-free) & Text Prompt & {\color{green}\checkmark} & {\color{green}\checkmark} & Multi-Stage \\
SkyFall-GS~\cite{skyfallgs} & Images & Multi-view Satellite & {\color{green}\checkmark} & {\color{green}\checkmark} & Iterative Optimization \\
Orbit2Ground~\cite{orbit2ground} & Images & Multi-view Satellite & {\color{green}\checkmark} & {\color{green}\checkmark} & Iterative Optimization \\
XCube~\cite{ren2024xcube} & Mesh & LiDAR Scan & {\color{green}\checkmark} & {\color{green}\checkmark} & 3D Generative \\
EarthCrafter~\cite{earthcrafter} & Mesh & Satellite + Depth & {\color{green}\checkmark} & {\color{green}\checkmark} & 3D Generative \\
\midrule
\textbf{Ours} & \textbf{3DGS} & \textbf{Single Satellite-View Image} & \textbf{\textcolor{green}{\checkmark}} & \textbf{\textcolor{green}{\checkmark}} & \textbf{3D Generative} \\
\bottomrule
\end{tabular}%
}
\end{table*}

%% file: body/2.related_work.tex
\section{Related Work}
\subsection{3D Object Generation}

Existing 3D object generation methods typically lift 2D priors into 3D via score distillation or view-conditioned image synthesis~\cite{poole2022dreamfusion, lin2023magic3d,wang2023prolificdreamer,liu2023zero}. 
Subsequent feed-forward and latent 3D frameworks improve efficiency~\cite{hong2024lrm,li2024instant3d,tang2024lgm,zhang2024gs,wei2024meshlrm}, but they remain largely bounded by object-centric representations.
More recently, latent 3D generative models~\cite{xiang2025structured,xiang2025nativecompactstructuredlatents,zhang2024clay,hong20243dtopia,wu2024direct3d,wang2023rodin} decouple 3D compression from generative modeling to enable high-quality object-scale synthesis. Our work follows this paradigm but develops and evaluates a structured latent representation for pre-optimized 3DGS at the scale of real-world aerial reconstructions, where the globally fixed primitive budgets common in object-level models become restrictive.

\subsection{3D Outdoor Scene Generation}
Existing outdoor scene creation systems synthesize training data, construct controllable urban assets, or reconstruct scenes from aerial observations~\cite{sat2scene,Qian_2026_Sat2Densitypp,qian2026satdgen,urbanworld,engstler2025syncity,CityDreamer,Qian_2023_Sat2Density,DBLP:conf/siggraph/ParishM01,10.1145/1141911.1141931}. Sat2City~\cite{sat2city} learns cascaded sparse-voxel latent diffusion from synthetic city assets, while Sat2City v2~\cite{hua2026sat2cityv2native3d} adapts a pretrained native structured latent model to real satellite--mesh pairs and generates textured mesh assets. Their goals and output representations differ from ours: few learn a generative model directly over pre-optimized real-world 3DGS reconstructions. XCube~\cite{ren2024xcube} and EarthCrafter~\cite{earthcrafter} advance learned 3D generation at larger spatial scales, with EarthCrafter expanding scenes through semantic-conditioned sliding-window inference. ABot-Earth 0.5~\cite{qian2026abotearth05generative3d} is an engineering-oriented project for satellite-conditioned 3DGS generation, emphasizing system integration, multi-LOD output, and planetary-scale deployment. Our focus is instead GS-Voxel itself: we describe how it converts pre-optimized 3DGS reconstructions into sparse spatial latents without per-scene fitting and how these latents are used for image-conditioned generation.

\subsection{Generative Modeling of 3D Gaussians}

Recent work has begun to explore diffusion-based generation for 3DGS. Early approaches mainly adopt view-conditioned or view-aligned Gaussian generation. For example, DiffusionGS~\cite{cai2025diffusiongs} relies on multi-view conditioning, while teacher-guided approaches~\cite{peng2025lessoninsplats} distill priors from 2D generative models. DiffSplat~\cite{lin2025diffsplat} and DiffGS~\cite{zhou2024diffgs} further introduce image-diffusion or functional parameterizations for Gaussian generation.

A particularly relevant line of work structures irregular Gaussian sets before generative modeling. Can3Tok~\cite{gao2025can3tok} uses cross-attention from input 3DGS primitives to learned canonical queries. Its preprocessing caps each reconstruction at 100K Gaussians and then selects 40K Gaussians for every VAE input; the VAE maps this globally fixed-cardinality set to a $64{\times}64{\times}4$ latent and decodes a fixed number of Gaussians. Can3Tok therefore demonstrates scene-level 3DGS latent modeling under a global primitive budget. GS-Voxel instead fixes capacity only within each active voxel, so the available output slots vary with the number of active voxels rather than being normalized to a common scene-wide count. This distinction matters in our evaluated aerial setting, where raw aerial 3DGS tiles can contain more than 3.0 million primitives.

GaussianCube~\cite{zhang2024gaussiancube} first obtains a fixed number of Gaussians through densification-constrained fitting and then rearranges them into a predefined voxel grid using Optimal Transport. L3DG~\cite{roessle2024l3dg} learns a sparse VQ-VAE over 3D Gaussians and performs latent diffusion for object- and room-scale synthesis.

TripoSplat is particularly relevant because its Density-Sampled Gaussian VAE (DeG-VAE) supports adaptive output budgets~\cite{yan2026triposplat}. Its encoder does not ingest an optimized 3DGS parameter set; instead, it samples points from an asset surface, projects DINOv3 and FLUX.2 VAE features from multi-view renderings onto those points, and encodes the resulting feature-augmented point sets. The two methods therefore assume different inputs and construct Gaussians differently: TripoSplat learns a new object-level Gaussian distribution from surface and image evidence, whereas GS-Voxel deterministically voxelizes an existing pre-optimized 3DGS set using a fixed local slot budget. TripoSplat reports decoded budgets from 33K to 262K Gaussians; in our evaluated aerial setting, raw $200\,\mathrm{m}\times200\,\mathrm{m}$ aerial 3DGS tiles can contain more than 3.0 million primitives, while the default GS-Voxel decoder provides approximately 1.5 million output slots. The two methods therefore operate at different scales and allocate output capacity differently: GS-Voxel's total slot capacity grows with the number of active voxels, but unlike TripoSplat, it does not adapt decoded density at inference time.

%% file: body/3.method.tex
\section{Method}
\label{sec:method}

Our method centers on GS-Voxel, a sparse voxel-aligned representation constructed deterministically from a compatible pre-optimized 3DGS reconstruction. We develop and evaluate this representation for large-scale aerial scenes. The full representation pipeline combines GS-Voxel construction with a factorized VAE that separately encodes voxel geometry and local Gaussian attributes.

\subsection{Fitting-Free Structured Latent Pipeline}

As illustrated in \cref{fig:overview}, the representation pipeline has two components. First, an explicit conversion reorganizes a compatible pre-optimized 3DGS reconstruction into sparse active voxels without additional per-scene optimization, a fixed-cardinality scene-wide Gaussian template, or global rearrangement of the input primitives. Second, a two-stage VAE maps voxel geometry and Gaussian attributes to compact sparse 3D latents. The resulting latents remain aligned with the voxel grid and can represent aerial scenes with millions of unevenly distributed primitives.

\input{body/fig_tex/overview}

\subsubsection{GS-Voxel Construction from Pre-optimized 3DGS}
\label{GS-voxel}

GS-Voxel converts the variable number of Gaussians in each voxel into a fixed-width feature vector and normalizes every attribute to a common range.

A compatible pre-optimized 3DGS reconstruction is represented as an unordered set of Gaussian primitives
\begin{equation}
\mathcal{G}=\{g_i\}_{i=1}^{N}, \qquad
g_i=\left(\mathbf{x}_i,\alpha_i,\mathbf{f}_i^{dc},\mathbf{s}_i,\mathbf{q}_i\right),
\end{equation}
where $\mathbf{x}_i\in\mathbb{R}^3$ is the primitive center, $\alpha_i$ is the opacity logit, $\mathbf{f}_i^{dc}\in\mathbb{R}^3$ is the DC spherical-harmonic color coefficient, $\mathbf{s}_i\in\mathbb{R}^3$ is the log-scale, and $\mathbf{q}_i\in\mathbb{R}^4$ is the rotation quaternion. In the current implementation, a compatible input follows this parameterization, is provided in a known cubic normalization frame, and stores only SH degree $0$, i.e., the DC color coefficient. SH0 gives view-independent color, which is suitable for our restricted aerial viewing range and keeps the per-primitive attribute vector small for multi-million-primitive scenes. The voxel-assignment rule itself does not depend on SH degree. Supporting higher-order view-dependent appearance would require adding the corresponding attribute channels and retraining the Local Attribute VAE and attribute flow.

Independently of the SH degree, native 3DGS is an unordered set, and different spatial regions contain different numbers of primitives. Sparse-convolutional latent models instead expect features attached to discrete spatial coordinates with a fixed channel dimension, so native 3DGS cannot be passed to them directly.

We therefore introduce \emph{GS-Voxel}, a sparse voxel-aligned representation that reorganizes Gaussian primitives into a format compatible with sparse 3D VAEs. Inspired by the practical offline conversion in O-Voxel~\cite{xiang2025nativecompactstructuredlatents}, we adopt a lightweight, fitting-free conversion without re-fitting or optimization. Unlike O-Voxel, which explicitly encodes surface geometry and material fields, GS-Voxel packs the retained primitives' local parameters directly (details in \cref{fig:gsvoxel}).

\input{body/fig_tex/gsvoxel}

We place the normalized scene in a cubic frame $[\mathbf{b}_{\min},\mathbf{b}_{\min}+Rh]^3$ and discretize it into a voxel grid of resolution $R^3$ with voxel size $h$. After discarding primitives below the implementation opacity threshold, each remaining Gaussian is assigned to a voxel according to its center,
\begin{equation}
\mathbf{v}_i=
\mathrm{clip}\!\left(
\left\lfloor \frac{\mathbf{x}_i-\mathbf{b}_{\min}}{h} \right\rfloor,\,
0,\,
R-1
\right),
\end{equation}
where $\mathbf{b}_{\min}$ is the minimum corner of the cubic frame. Let $\mathcal{G}(v)$ denote the set of retained, above-threshold primitives whose centers fall into voxel $v$. Since the number of Gaussians within each voxel is variable, we convert the local unordered set into a fixed-capacity representation by sorting primitives by their opacity logits $\alpha$ and retaining the top-$K_{\mathrm{in}}$ entries:
\begin{equation}
\tilde{\mathcal{G}}(v)=\mathrm{TopK}_{\alpha}\!\left(\mathcal{G}(v),K_{\mathrm{in}}\right),
\qquad
n_v=\min\!\left(|\mathcal{G}(v)|,K_{\mathrm{in}}\right).
\end{equation}
The remaining slots are zero-padded when $n_v<K_{\mathrm{in}}$. This yields a deterministic local ordering and, more importantly, ensures a consistent channel dimension across active voxels for downstream neural compression and generative modeling. In practice, most occupied voxels contain only a few primitives, so a proper choice of $K_{\mathrm{in}}$ preserves local content with controlled capacity while avoiding unnecessarily large feature channels.

For each retained Gaussian, we encode its position relative to the voxel center $\mathbf{m}_v$,
\begin{equation}
\Delta \mathbf{x}_{v,j}=\frac{\mathbf{x}_{v,j}-\mathbf{m}_v}{h},
\end{equation}
which retains sub-voxel offsets before 8-bit quantization. We normalize each retained Gaussian's local position and rendering attributes to $[0,1]$ before quantizing them into uint8 buffers:
\begin{align}
\mathbf{o}_{v,j} &= \Delta \mathbf{x}_{v,j}+0.5, \\
\mathbf{c}_{v,j} &= C_0\mathbf{f}^{dc}_{v,j}+0.5, \\
p_{v,j} &= \sigma(\alpha_{v,j}), \\
\mathbf{l}_{v,j} &=
\frac{\mathrm{clip}(\mathbf{s}_{v,j},s_{\min},s_{\max})-s_{\min}}{s_{\max}-s_{\min}}, \\
\mathbf{r}_{v,j} &= \frac{\hat{\mathbf{q}}_{v,j}+1}{2},
\end{align}
where $C_0=0.28209479$, $\hat{\mathbf{q}}_{v,j}$ is the normalized quaternion, and $(s_{\min}, s_{\max}) = (-24,-4)$ in our implementation. The final GS-Voxel feature stored at each active voxel is
\begin{equation}
\phi_v=\mathrm{concat}\!\left[
\mathrm{vec}(\mathbf{O}_v),\,
\mathrm{vec}(\mathbf{C}_v),\,
\mathrm{vec}(\mathbf{P}_v),\,
\mathrm{vec}(\mathbf{L}_v),\,
\mathrm{vec}(\mathbf{R}_v)
\right].
\end{equation}
Here, $n_v$ is only a construction-time count used for truncation and zero padding: it is not included in $\phi_v$, supplied to the VAE, or predicted by the decoder. Active voxels are those with at least one retained, above-threshold primitive, and zero-padded slots have zero opacity and do not contribute to rendering.

The conversion from pre-optimized 3DGS to GS-Voxel is explicit and lightweight: it voxelizes the primitives, retains up to $K_{\mathrm{in}}$ dominant entries per active voxel, and packs their normalized attributes into a sparse format. The inverse mapping dequantizes the active-voxel features, reconstructs local offsets and Gaussian parameters, and concatenates the resulting primitives over the active support. Complete forward and inverse procedures, default conversion parameters, and serialization details are provided in Sec.~\ref{sec:supp_conversion} of the supplementary material. Under the tested choices of voxel size and $K_{\mathrm{in}}$, this explicit conversion retains high rendering fidelity, as evaluated in our experiments. GS-Voxel therefore retains the selected local primitives in a sparse voxel grid with the fixed per-voxel feature size required by latent models. We denote its fine active support at resolution $R^3$ by $\mathcal{S}_R$.

\subsubsection{Two-Stage GS-Voxel VAE}

We implement the GS-Voxel factorized VAE as two components: a \emph{Geometry VAE}, which models voxel geometry as hierarchical subdivision and occupancy, and a \emph{Local Attribute VAE}, which models Gaussian primitives only within occupied voxels. This decomposition separates support modeling from attribute reconstruction and avoids allocating attribute capacity to empty space. These two GS-Voxel-specific components are distinct from the Sparse Structure VAE used for the coarse stage of the generation pipeline.

\input{body/fig_tex/infer}

\paragraph{Geometry VAE.}
The Geometry VAE encodes the fine support $\mathcal{S}_R$ into a geometry latent $z_g$ on a coarsened sparse lattice at $(R/8)^3$. Each active coordinate is augmented with a sinusoidal positional encoding before entering the encoder. The decoder predicts hierarchical child-subdivision signals whose decisions define the reconstructed fine support $\hat{\mathcal{S}}_R$:
\begin{equation}
\{\hat{s}_l\}_{l=1}^{L} = D_g(z_g).
\end{equation}
where $L$ is the number of hierarchy levels and $s_l$ is the target child-occupancy signal at level $l$. Unlike TRELLIS.2, which also models additional geometric quantities such as dual vertices, our geometry stage is simplified for GS-Voxel and only reconstructs occupied support and its hierarchy. The training objective is
\begin{equation}
\mathcal{L}_{\mathrm{geom}}
= \lambda_{\mathrm{subdiv}}\sum_{l=1}^{L}\mathrm{BCE}(\hat{s}_l,s_l)
+ \lambda_{\mathrm{KL}}\mathcal{L}_{\mathrm{KL}}.
\end{equation}

\paragraph{Local Attribute VAE.}
The Local Attribute VAE encodes Gaussian attributes defined on the fine GS-Voxel support; it does not learn the support itself. During generation, the attribute flow produces $z_a$, and the Local Attribute VAE decoder maps it to Gaussian attributes on the generated $\hat{\mathcal{S}}_R$.

For an active voxel $v$, the packed VAE input is
\begin{equation}
\begin{aligned}
\mathcal{U}(v) &= \{\mathbf{u}_{v,j}\}_{j=1}^{K_{\mathrm{in}}},\\
\mathbf{u}_{v,j} &=
(\mathbf{o}_{v,j},\mathbf{c}_{v,j},p_{v,j},
\mathbf{l}_{v,j},\mathbf{r}_{v,j}) \in [0,1]^{14},
\end{aligned}
\end{equation}
where the components are the normalized quantities defined in Eqs.~(5)--(9). The encoder maps these fine-grid attributes to a compact latent $z_a$ on a coarsened sparse lattice at $(R/8)^3$. During decoding, rather than predicting a variable number of primitives within each supplied fine-grid voxel, we adopt a fixed per-voxel slot decoder that outputs
\begin{equation}
\hat{\mathcal{U}}(v)=\{\hat{\mathbf{u}}_{v,j}\}_{j=1}^{K_{\mathrm{out}}},
\end{equation}
for each voxel on the supplied fine support. We apply the inverse transforms of Eqs.~(5)--(9) to reconstruct sub-voxel offsets and standard 3DGS attributes before differentiable rendering. Importantly, $K_{\mathrm{out}}$ need not equal $K_{\mathrm{in}}$: supervision is applied through rendered-image consistency rather than explicit count matching, so the decoder may use a different local slot budget as long as the rendered 3DGS matches the target observations.

The attribute loss is
\begin{equation}
\mathcal{L}_{\mathrm{attr}}
= \lambda_{\mathrm{img}}\mathcal{L}_{\mathrm{img}}
+ \lambda_{\mathrm{KL}}\mathcal{L}_{\mathrm{KL}},
\end{equation}
where $\mathcal{L}_{\mathrm{img}}$ is a rendering loss combining L1, L2, SSIM, and perceptual terms. Therefore, this stage does not reconstruct sparse support or predict an explicit primitive count; it learns fixed per-voxel Gaussian slots on a supplied fine support.

\subsection{Generation Pipeline}

For image-conditioned generation, we use a three-stage flow-matching pipeline based on TRELLIS.2~\cite{xiang2025nativecompactstructuredlatents}. The sparse-structure flow models a coarse occupancy latent, while the geometry and attribute flows model the two GS-Voxel-specific latents.

The generation process has three stages. First, the image-conditioned sparse-structure flow predicts a coarse occupancy latent, which the Sparse Structure VAE decoder converts into support at $(R/8)^3$. Second, conditioned on the input image and this coarse support, the geometry flow predicts $z_g$; the Geometry VAE decoder expands it into the hierarchical subdivision and fine active support. Third, conditioned on the input image and generated geometry, the attribute flow predicts $z_a$; the Local Attribute VAE decoder emits $K_{\mathrm{out}}$ Gaussian slots per active fine-grid voxel. The resulting parameters form a standard 3DGS primitive set.

\subsubsection{Large-Area Scene Generation}
For large-area synthesis beyond a fixed training crop, we use an overlap-aware tiled-inference procedure tailored to our three-stage latent hierarchy. Existing neighboring tiles provide inpainting constraints, while stage-wise structure, geometry, and attribute generation together with overlap blending maintain continuity across tile boundaries. At inference, each spatial tile is conditioned on the aligned patch of a supplied satellite-view image, which may be real or generated. Implementation details are provided in the supplemental material.

%% file: body/fig_tex/overview.tex
\begin{figure}[htp]
  \centering
  \includegraphics[width=\linewidth]{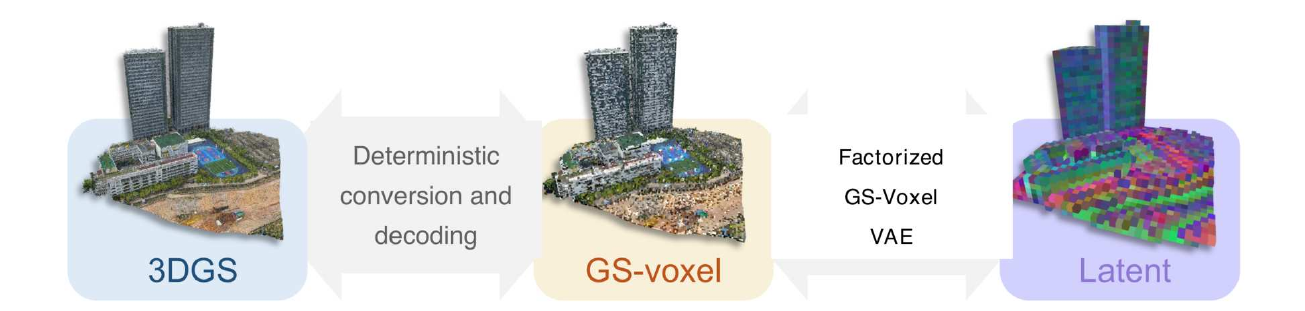}
  \Description{}
    \caption{Overview of our structured-latent pipeline. We first convert a compatible pre-optimized 3DGS reconstruction into GS-Voxel, a sparse representation of the retained local Gaussian parameters, and then encode it with a factorized VAE. The Geometry VAE models hierarchical subdivision and occupancy, while the Local Attribute VAE models Gaussian attributes on the decoded support.}
  \label{fig:overview}
\end{figure}

%% file: body/fig_tex/gsvoxel.tex
\begin{figure}[htp]
  \centering
  \includegraphics[width=\linewidth]{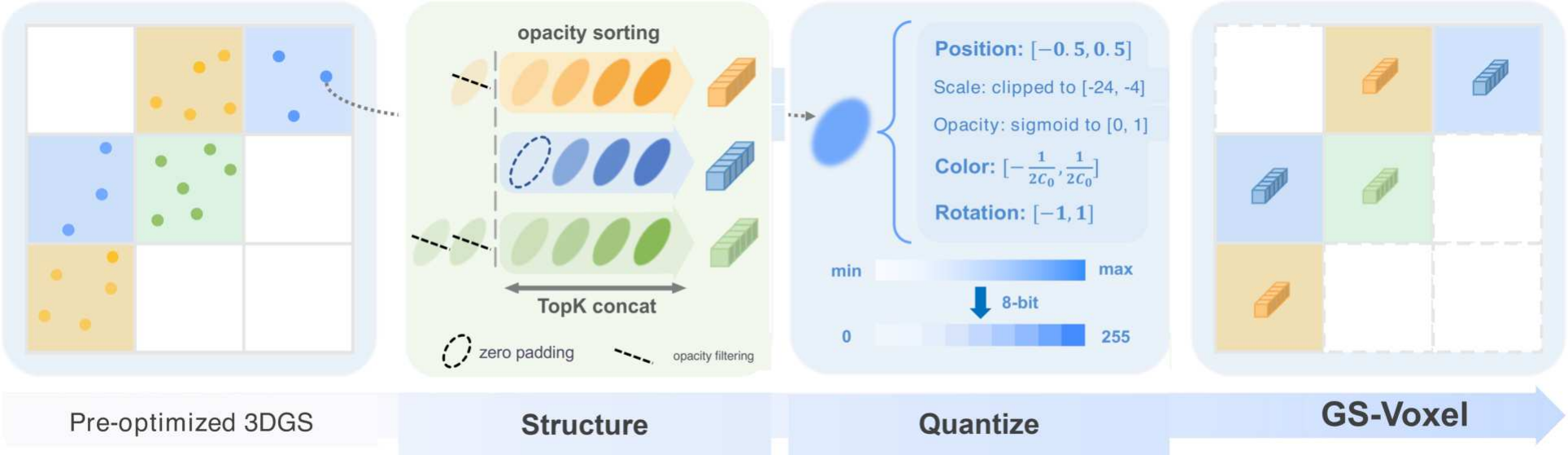}
  \Description{}
    \caption{Deterministic conversion from a compatible pre-optimized 3DGS reconstruction to GS-Voxel.}
  \label{fig:gsvoxel}
  
\end{figure}

%% file: body/fig_tex/infer.tex
\begin{figure*}[t]
  \centering
  \includegraphics[width=\linewidth]{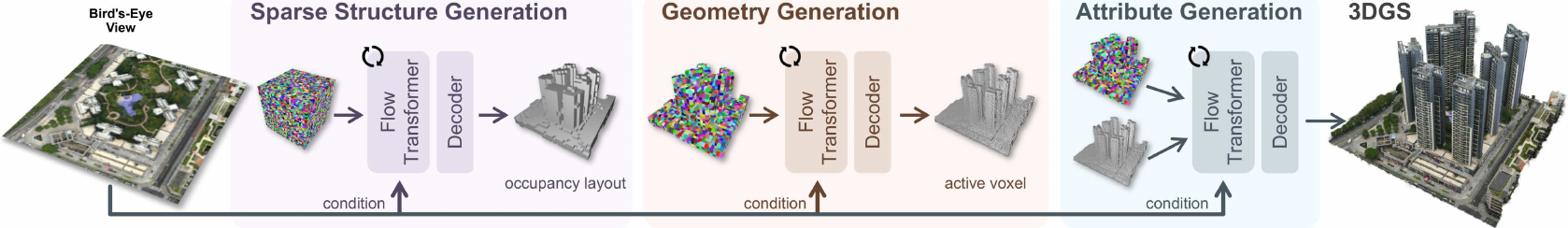}
  \Description{}
    \caption{Conditional generation pipeline. Given a satellite-view image, a coarse sparse-structure stage predicts scene support; GS-Voxel-specific geometry and attribute stages then generate fine support and Gaussian parameters, yielding a standard 3DGS primitive set.}
  \label{fig:infer}
\end{figure*}

%% file: body/4.results.tex
\section{Training Data Construction}
\label{sec.dataprocess}
We construct the dataset from multiple large-scale aerial 3DGS scenes. Each source scene spans hundreds of meters, has uneven terrain, and may contain floating outliers. Preprocessing proceeds in three stages. We first crop each scene into fixed-size windows, normalize each window to a unit cube, remove outliers, and convert the remaining Gaussians into GS-Voxel. We then sample a dense set of virtual cameras to render diverse aerial views. Finally, geometric and VLM-based filters select reliable renders to supervise the Local Attribute VAE. Details of the three stages follow.

\paragraph{Training Tile Generation and Voxel Filtering}
We perform sliding-window cropping on the \emph{large-scale} 3DGS for each scene.
Specifically, we slide an axis-aligned window of size $W=\SI{200}{\meter}$ with a stride $S=\SI{150}{\meter}$ ($\SI{50}{\meter}$ overlap) across the large 3DGS, and retain only the Gaussians whose centers fall inside each window footprint. Each cropped window is then normalized to the unit cube $[-0.5,0.5]^3$. 
We apply DBSCAN ($\varepsilon{=}0.02$, $m_\text{min}{=}50$) to isolate the dominant scene support from floating outliers and retain its largest cluster. We then use the lower extent of this retained cluster as the vertical reference and translate it so that its lowest point lies at $z=-0.5$, aligning the scene support with the lower face of the unit cube. This ordering removes spatially disconnected primitives before they can create spurious voxels and prevents bottom outliers from determining the vertical alignment.
Finally, following Sec.~\ref{GS-voxel}, the cleaned Gaussians are converted into \emph{GS-Voxel} for training.

\paragraph{Virtual Camera Setting}
After normalization, we render a dense set of training views in the normalized coordinate frame. We place a virtual aerial camera rig on a $10{\times}10$ grid spanning $[-0.7,0.7]^2$ in the $xy$ plane. The grid extends slightly beyond the scene footprint so that peripheral cameras can also capture the scene at oblique angles, and is replicated across five height layers with denser sampling and wider FOV near the ground. Here pitch is measured as the deviation from the nadir direction, so $\text{pitch}{=}0^\circ$ denotes a top-down view; for non-zero pitch we additionally sweep four compass yaws so that every grid cell is observed from multiple orientations. Camera orientation is set directly from each layer's pitch and compass yaw rather than by looking at the scene center, keeping view distributions comparable across windows of different content. To break the regularity of the rig and increase view diversity, we apply uniform random perturbations to the $xy$ position ($\pm 15\%$ of the grid spacing), height ($\pm 0.1$), pitch ($\pm 5^\circ$), and yaw ($\pm 30^\circ$). The per-layer configuration is summarized in Table~\ref{tab:cam-layers}. The rig produces approximately $3.9$K candidate views per window, which are then filtered as described below.

\input{body/table_tex/camera}

\paragraph{VLM-Based View Filtering}
For each window, the camera rig generates about 3,900 candidate renders. Occlusions, blind spots, and reconstruction artifacts make many views unreliable supervision (e.g., empty boundary views, oblique views through Gaussian holes, or renders with blur and floaters). We therefore apply two sequential filters: (1) a geometric coverage test using rendered alpha, discarding views with mean accumulated opacity below $\tau_\alpha=0.9$; and (2) VLM-based quality scoring via Qwen-VL-Plus from the Qwen-VL model family~\cite{qwenvl}, which evaluates the remaining renders in $[0,1]$ for texture and silhouette sharpness, reconstruction artifacts, and whether dark regions correspond to natural boundaries. The exact scoring prompt is provided in the supplementary material. The filtering policy retains roughly 500--1,000 reliable renders per window. These views supervise only the Local Attribute VAE and are separate from the conditioning-view pool used by the flow models.

\input{body/fig_tex/data_quality}

\section{Experiments}

\subsection{Implementation Details}
Following Sec.~\ref{sec.dataprocess}, we obtain approximately 18K samples, each covering a $200 \times 200~\mathrm{m}^2$ area. We use 800 samples for validation and the rest for training. The split is performed at the source-scene level: no source scene or spatial region is shared between training and validation.

The two GS-Voxel-specific VAEs use three spatial downsampling stages, taking inputs at $R=256$ and producing latents at resolution $32^3$. The pretrained Sparse Structure VAE is reused without retraining, while the sparse-structure, geometry, and attribute conditional flows are all trained on our data. For this training, each 3D tile is paired with nine $512 \times 512$ near-orthographic bird's-eye-view renders with slightly perturbed extrinsics. The same nine-view pool is used by all three conditional flows, with one view randomly sampled for each training instance.

We train the Geometry VAE, Local Attribute VAE, and all three conditional flows with AdamW~\cite{loshchilov2018decoupled}, using a learning rate of $1\times10^{-4}$ and weight decay $0.01$. For the conditional flows, we use classifier-free conditioning dropout with a rate of $0.1$. We use 0.3B-parameter DiTs for the geometry and attribute flows, reduced from the 1.3B architecture in TRELLIS.2 to match our smaller and less diverse dataset.

\subsection{Representation and VAE Analysis}

We evaluate the representation and VAE analyses in this subsection on 10 validation tiles from the source-scene-separated split. Reported reconstruction metrics are aggregated over rendered views from these tiles.

\paragraph{Choice of $K_{\mathrm{in}}$ and $R$ for GS-Voxel Construction.}
Table~\ref{tab:ablation_kin_r} evaluates the direct GS-Voxel conversion, before VAE encoding, while varying the per-voxel budget $K_{\mathrm{in}}$ and grid resolution $R$. At fixed $R=256$, increasing $K_{\mathrm{in}}$ improves reconstruction because more local primitives are retained. The gain becomes small from $K_{\mathrm{in}}=16$ to $32$, while the per-voxel attribute width doubles from $224$ to $448$. At fixed $K_{\mathrm{in}}=4$, increasing $R$ from $256$ to $512$ raises PSNR from 26.82 to 39.20, but also increases the average active support from approximately 377K to 980K voxels. The tested $R=512$, $K_{\mathrm{in}}=4$ setting remains slightly below the fidelity of $R=256$, $K_{\mathrm{in}}=16$ while using about $2.6\times$ as many active voxels. Because each tile covers approximately $200\,\mathrm{m}\times200\,\mathrm{m}$, we adopt $R=256$ and $K_{\mathrm{in}}=16$ as the tested trade-off between direct-conversion fidelity and sparse-support size.

\input{body/table_tex/ablation_for_gs_voxel}

\paragraph{Choice of $K_{\mathrm{out}}$ for Local Attribute VAE.}
We ablate the per-voxel output slot budget $K_{\mathrm{out}}$ of the Local Attribute VAE over $\{1,4,8\}$. As shown in Table~\ref{tab:kout_attr}, $K_{\mathrm{out}}=1$ restricts local capacity and lowers reconstruction quality. Although $K_{\mathrm{out}}=8$ provides more slots, it does not improve reconstruction under the current objective. We therefore use $K_{\mathrm{out}}=4$, which gives the best PSNR, SSIM, and LPIPS among the tested settings.

\input{body/table_tex/kout}

\paragraph{Factorized VAE Component Analysis.}
We compare our factorized VAE with a single-stage baseline that jointly compresses voxel geometry and local Gaussian attributes. The Geometry VAE and Local Attribute VAE are evaluated separately because they model different quantities. As shown in Table~\ref{tab:two_stage_vs_single}, the Geometry VAE obtains 0.99 Shape IoU for support reconstruction. The Local Attribute VAE is evaluated on the target support $\mathcal{S}_R$ and obtains 23.09 PSNR, 0.62 SSIM, and 0.331 LPIPS.

\input{body/table_tex/twostagecompare}

\subsection{Tile-Level Generative Results}
For context, \cref{tab:conditions_results} lists cross-paper FID and KID values reported by existing methods, whose ground-truth datasets and camera protocols differ. Under our evaluation protocol, the generated and ground-truth sets each contain 15K images, with ground truth rendered from held-out source scenes in the validation split; our method obtains an FID of 28.0 and a KID of 0.020. Qualitative tile-level generation results are shown in \cref{fig:tile_result}.

\input{body/table_tex/tab_fid}

\subsection{Large-Area Scene Generation}

As shown in \cref{teaser,fig:large_scale_result1,fig:large_scale_result2}, we demonstrate large-area synthesis using constructed satellite-view conditions whose source scenes are geographically disjoint from the training data. We assemble these test conditions from rendered aerial references by spatially extending local content at a fixed ground sampling resolution. Overlap-aware tiled inference generates scene latents across the full footprint and decodes them into standard 3DGS primitives. The examples span up to $1{,}400\,\mathrm{m}\times800\,\mathrm{m}$, substantially beyond the $200\,\mathrm{m}\times200\,\mathrm{m}$ training crop.

%% file: body/table_tex/camera.tex
\begin{table}[t]
\caption{Per-layer configuration of the virtual aerial camera rig. Pitch is measured as the deviation from the nadir direction. For $\text{pitch}{=}0^\circ$ a single yaw is used (top-down view); for $\text{pitch}{>}0^\circ$ four compass yaws $\{0^\circ,90^\circ,180^\circ,270^\circ\}$ are sampled. Views/cell counts cameras per $xy$ grid cell; Views/window assumes a $10{\times}10$ grid.}
\vspace{-0.5em}
\centering
\setlength{\tabcolsep}{6pt}
    \resizebox{\linewidth}{!}{
\begin{tabular}{c c c l c c}
\toprule
Layer & $z$ & FOV & Pitches (deg) & Views/cell & Views/window\\
\midrule
0 & $0.0$ & $26^\circ$ & $\{0, 30, 45, 60\}$  & $13$ & $1{,}300$\\
1 & $0.3$ & $18^\circ$ & $\{15, 30, 45\}$     & $12$ & $1{,}200$\\
2 & $0.6$ & $14^\circ$ & $\{0, 15, 30\}$      & $9$  & $\phantom{0}900$\\
3 & $0.9$ & $14^\circ$ & $\{15\}$             & $4$  & $\phantom{0}400$\\
4 & $1.2$ & $14^\circ$ & $\{0\}$              & $1$  & $\phantom{0}100$\\
\midrule
\multicolumn{4}{r}{Total} & $39$ & $3{,}900$\\
\bottomrule
\end{tabular}
}
\label{tab:cam-layers}
\end{table}

%% file: body/fig_tex/data_quality.tex
\begin{figure}[t]
  \centering
  \includegraphics[width=\linewidth]{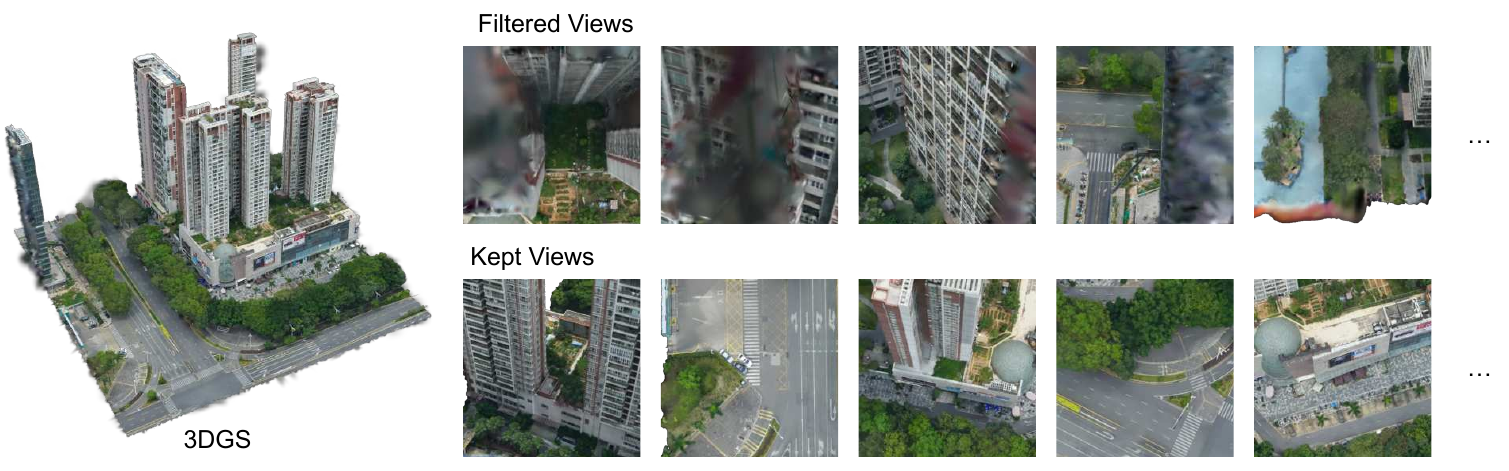}
  \Description{Image Filtering Results}
  \caption{Examples of retained and rejected aerial renders. Filtering removes unreliable or severely degraded supervision; only the retained views supervise the Local Attribute VAE.
}
  \label{fig:qwen_filter_examples}
\end{figure}

%% file: body/table_tex/ablation_for_gs_voxel.tex
\begin{table}[t]
    \centering
    \caption{Direct GS-Voxel conversion before VAE encoding, evaluated on 10 validation tiles. We report average retained-primitives and active-voxel counts rounded to the nearest thousand (K), together with rendered reconstruction quality. Bold settings denote our default configuration.}
  \vspace{-0.5em}
    \label{tab:ablation_kin_r}
    \resizebox{\linewidth}{!}{
    \begin{tabular}{c c c c c c c}
    \toprule
    $R$ & $K_{\mathrm{in}}$ & Avg. Retained Primitives & Avg. Active Voxels & SSIM $\uparrow$ & PSNR $\uparrow$ & LPIPS $\downarrow$ \\
    \midrule
    256 & 1  & 377K   & 377K & 0.55 & 18.52 & 0.425 \\
    256 & 4  & 971K   & 377K & 0.85 & 26.82 & 0.166 \\
    256 & 8  & 1{,}340K & 377K & 0.95 & 33.26 & 0.058 \\
    \textbf{256} & \textbf{16} & 1{,}602K & 377K & 0.98 & 40.04 & 0.023 \\
    256 & 32 & 1{,}737K & 377K & 0.98 & 40.94 & 0.021 \\
    512 & 1  & 980K   & 980K & 0.87 & 27.38 & 0.156 \\
    512 & 4  & 1{,}501K & 980K & 0.98 & 39.20 & 0.025 \\
    \bottomrule
    \end{tabular}
    }
  \end{table}

%% file: body/table_tex/kout.tex
\begin{table}[h]
\centering
\caption{Choice of $K_{\mathrm{out}}$ for the Local Attribute VAE, evaluated using rendered-image reconstruction quality.}
\vspace{-0.5em}
\label{tab:kout_attr}
\resizebox{\linewidth}{!}{%
\begin{tabular}{l c c c}
\toprule
Method / Setting & PSNR $\uparrow$ & SSIM $\uparrow$ & LPIPS $\downarrow$ \\
\midrule
$K_{\mathrm{out}}=1$ & 22.81 & 0.61 & 0.350 \\
$K_{\mathrm{out}}=4$ (Ours) & \textbf{23.09} & \textbf{0.62} & \textbf{0.331} \\
$K_{\mathrm{out}}=8$ & 22.12 & 0.58 & 0.354 \\
\bottomrule
\end{tabular}
}
\end{table}

%% file: body/table_tex/twostagecompare.tex
\begin{table}[t]
\centering
\caption{Separate evaluation of the Geometry VAE and Local Attribute VAE. Shape IoU measures support reconstruction; PSNR, SSIM, and LPIPS measure attribute reconstruction on the target support.}
  \vspace{-1em}
\label{tab:two_stage_vs_single}
\resizebox{\linewidth}{!}{
\begin{tabular}{lcccc}
\toprule
Model & Shape IoU $\uparrow$ & PSNR $\uparrow$ & SSIM $\uparrow$ & LPIPS $\downarrow$ \\
\midrule
Single-stage VAE          & 0.76 & 21.01 & 0.54 & 0.418 \\
Geometry VAE              & \textbf{0.99} & -- & -- & -- \\
Local Attribute VAE       & -- & \textbf{23.09} & \textbf{0.62} & \textbf{0.331} \\
\bottomrule
\end{tabular}
}
\end{table}

%% file: body/table_tex/tab_fid.tex
\begin{table}[t]
\caption{Cross-paper reference for image-level FID/KID. Baselines use different ground-truth sets and camera protocols; values are not directly comparable and no ranking is implied.}
\centering
\small
\setlength{\tabcolsep}{10pt}
\begin{tabular}{lcc}
\toprule
\textbf{Method} & \textbf{FID} & \textbf{KID} \\
\midrule
CityDreamer~\cite{CityDreamer} & 97.3 & 0.096 \\
GaussianCity~\cite{Xie_2025_CVPR} & 86.9 & 0.090 \\
EarthCrafter~\cite{earthcrafter} & 69.5 & 0.061 \\
\hline
Ours & 28.0 & 0.020 \\
\bottomrule
\end{tabular}
\label{tab:conditions_results}
\end{table}

%% file: body/5.conclusion.tex
\section{Conclusion}

We presented \ourmethod, a fitting-free structured latent representation, and instantiated it for large-scale aerial 3D Gaussian scene generation. Its deterministic construction organizes selected primitives from a compatible pre-optimized 3DGS reconstruction into sparse voxels, while its factorized VAE encodes occupied support separately from local Gaussian attributes. In our aerial experiments, this design handles millions of input primitives without fitting each scene to a fixed-size, scene-wide Gaussian template. We use image-conditioned flow models to generate sparse-structure, geometry, and attribute latents that decode into aerial 3DGS scenes, while overlap-aware tiled inference extends synthesis beyond a single training crop. GS-Voxel may therefore provide a basis for future work on scalable aerial 3D generation and downstream simulation.

%% file: body/6.limitations.tex
\section{Limitations and Future Work}

GS-Voxel is a representation and factorized VAE framework for compatible pre-optimized 3DGS reconstructions; the present implementation and experiments are developed for SH0 aerial scenes.

Because training relies on 3DGS scenes reconstructed from real-world data, the approach is constrained by the scale and availability of suitable training scenes. The current image-conditioned flow models are trained with rendered bird's-eye-view images. At inference, the conditional pipeline can accept real or generated satellite-view images, but robustness to sensor variation and domain shifts among rendered, real, and generated conditions has not yet been evaluated. Extending training and evaluation across these input domains is important future work.

GS-Voxel's top-$K_{\mathrm{in}}$ local capacity can discard primitives in extremely dense voxels, while finite spatial resolution can make very thin structures harder to preserve and reconstruct. Adaptive resolution and learned local-capacity allocation are promising directions for addressing these distinct limitations.

The present study evaluates reconstruction fidelity and image-conditioned generation rather than coding efficiency, and therefore does not yet report entropy-coded bitrates or rate--distortion curves. A systematic evaluation of memory and storage reduction, together with quantized or entropy-coded GS-Voxel latents, is planned for a future revision.

Beyond aerial scenes, future work will test the same fitting-free voxel-local representation on indoor scenes, street-level scenes, and object-level assets. These settings introduce different camera distributions, spatial scales, and appearance characteristics, and may require adapting the current SH0 attributes, aerial-scene normalization policy, and training data.

Finally, future work should evaluate whether generated aerial scenes are sufficiently faithful and controllable for downstream uses such as planning, simulation, or emergency-response analysis.

%% file: body/6.appendix.tex
\section{GS-Voxel Conversion Procedures}
\label{sec:supp_conversion}

The following pseudocode gives the complete deterministic conversion used to construct GS-Voxel and recover a standard 3DGS primitive set.

\paragraph{Input normalization and coordinate recovery.}
The conversion assumes that a selected cubic 3DGS crop has been placed in the normalized frame; it does not estimate this frame from the primitive bounds. Let $\mathbf{c}\in\mathbb{R}^3$ and $L>0$ be the center and side length of the crop in the original coordinate system. Before Algorithm~S1, we transform the Gaussian centers and log-scales as
\begin{equation}
\mathbf{x}_i=\frac{\mathbf{x}^{\mathrm{world}}_i-\mathbf{c}}{L},
\qquad
\mathbf{s}_i=\mathbf{s}^{\mathrm{world}}_i-\log L,
\label{eq:supp_world_to_normalized}
\end{equation}
where subtraction of $\log L$ is applied to all three log-scale components. This maps the crop to $[-0.5,0.5]^3$ while preserving each Gaussian's size relative to the scene. Opacity logits, SH coefficients, and rotations are unchanged. If recovery of the original coordinate system is required, preprocessing must retain $(\mathbf{c},L)$; these values are not stored in the \texttt{.vxz} file. Algorithm~S2 returns Gaussians in the normalized frame; world-space centers and log-scales are recovered by
\begin{equation}
\mathbf{x}^{\mathrm{world}}_i=L\mathbf{x}_i+\mathbf{c},
\qquad
\mathbf{s}^{\mathrm{world}}_i=\mathbf{s}_i+\log L.
\label{eq:supp_normalized_to_world}
\end{equation}

\paragraph{Default configuration.}
For all experiments, the input is an SH0 3DGS tile normalized to $[-0.5,0.5]^3$. We use $R=256$, voxel size $h=1/256$, local input capacity $K_{\mathrm{in}}=16$, opacity threshold $t_8=5$ (i.e., $5/255$ after sigmoid), 8-bit attribute quantization, log-scale range $[s_{\min},s_{\max}]=[-24,-4]$, and an out-of-bounds tolerance $\delta_b=2/128$. Centers lying within $\delta_b$ outside the normalized cube are clamped to boundary voxels; a larger deviation is treated as an invalid normalization. No scale-based pruning is used. The choices $R=256$ and $K_{\mathrm{in}}=16$ follow the fidelity--support trade-off evaluated in Table~\ref{tab:ablation_kin_r}; the remaining conversion settings are fixed across all experiments. A PLY input uses the standard fields \texttt{x/y/z}, \texttt{f\_dc\_0/1/2}, \texttt{opacity}, \texttt{scale\_0/1/2}, and \texttt{rot\_0/1/2/3}, where opacity and scale are stored as logits and log-scales, respectively.

\paragraph{Serialization.}
We serialize GS-Voxel as a \texttt{.vxz} file using the open-source \texttt{o\_voxel.io} interface released with TRELLIS.2, available at \url{https://github.com/microsoft/TRELLIS.2/tree/main/o-voxel}. O-Voxel is used only as the sparse serialization layer; the stored Gaussian feature schema is specific to GS-Voxel. For $M$ active voxels, the file contains coordinates of shape $M\times3$ and uint8 buffers \texttt{offset}, \texttt{rgb}, \texttt{opacity}, \texttt{scale}, and \texttt{rotation} with respective shapes $M\times3K_{\mathrm{in}}$, $M\times3K_{\mathrm{in}}$, $M\times K_{\mathrm{in}}$, $M\times3K_{\mathrm{in}}$, and $M\times4K_{\mathrm{in}}$. Our conversion utility also stores an auxiliary uint8 \texttt{count} buffer of shape $M\times1$ to identify the retained slots. This buffer is not part of $\phi_v$ and is not supplied to the VAE; without it, valid slots are equivalently identified by quantized opacity greater than or equal to $t_8$, because padded slots have zero opacity. All stored GS-Voxel data are contained in this single \texttt{.vxz} file.

Let
\begin{equation}
Q_8(y)=\left\lfloor 255\,\mathrm{clip}(y,0,1)\right\rfloor,
\qquad D_8(z)=z/255
\end{equation}
denote 8-bit quantization and dequantization. The opacity threshold is represented by an integer $t_8\in[0,255]$; we use $t_8=5$.

\begin{tcolorbox}[
  colback=black!1,
  colframe=black!45,
  boxrule=0.6pt,
  arc=2pt,
  left=2mm,right=2mm,top=1.5mm,bottom=1.5mm,
  title={Algorithm S1: 3DGS $\rightarrow$ GS-Voxel},
  fonttitle=\normalfont,
  breakable
]
Input: normalized SH0 Gaussian set $\mathcal{G}$; cubic frame $(\mathbf{b}_{\min},R,h)$; local capacity $K_{\mathrm{in}}$; opacity threshold $t_8$; log-scale range $[s_{\min},s_{\max}]$.\\
Output: sparse active coordinates and packed features $\mathcal{V}=\{(v,\phi_v)\}$.
\begin{enumerate}[leftmargin=5mm,itemsep=1pt,topsep=2pt]
  \item Verify the normalized spatial range. Clamp centers within $\delta_b$ of the cube boundary and stop if any center exceeds this tolerance.
  \item Compute $p_i=\sigma(\alpha_i)$ and discard every $g_i$ with $p_i\leq t_8/255$.
  \item Assign each remaining primitive to $\mathbf{v}_i$ with Eq.~(2), and compute the corresponding voxel center.
  \item Compute the normalized offset, color, opacity, scale, and rotation $(\mathbf{o}_i,\mathbf{c}_i,p_i,\mathbf{l}_i,\mathbf{r}_i)$ using Eqs.~(5)--(9).
  \item Group primitives by voxel coordinate. For every nonempty voxel $v$, sort its group by decreasing opacity and retain the first $K_{\mathrm{in}}$ primitives.
  \item Apply $Q_8$ independently to every normalized attribute of each retained primitive.
  \item Place the quantized attributes in opacity order, zero-pad unused local slots to $K_{\mathrm{in}}$, and concatenate the buffers into $\phi_v$ as in Eq.~(10).
  \item Return all active coordinates and their packed features.
\end{enumerate}
\end{tcolorbox}

\begin{tcolorbox}[
  colback=black!1,
  colframe=black!45,
  boxrule=0.6pt,
  arc=2pt,
  left=2mm,right=2mm,top=1.5mm,bottom=1.5mm,
  title={Algorithm S2: GS-Voxel $\rightarrow$ 3DGS},
  fonttitle=\normalfont,
  breakable
]
Input: sparse GS-Voxel $\mathcal{V}$; cubic frame $(\mathbf{b}_{\min},R,h)$; opacity threshold $t_8$; log-scale range $[s_{\min},s_{\max}]$.\\
Output: reconstructed Gaussian set $\hat{\mathcal{G}}$.
\begin{enumerate}[leftmargin=5mm,itemsep=1pt,topsep=2pt]
  \item Initialize $\hat{\mathcal{G}}\leftarrow\varnothing$.
  \item For each active voxel $(v,\phi_v)$, unpack $\phi_v$ into its $K_{\mathrm{in}}$ quantized local slots and discard slots whose quantized opacity is below $t_8$.
  \item For every remaining slot, apply $D_8$ to obtain $(\mathbf{o},\mathbf{c},p,\mathbf{l},\mathbf{r})$, set $\bar p=\mathrm{clip}(p,10^{-4},1-10^{-4})$, and reconstruct
  \begin{align*}
  \mathbf{x} &= \mathbf{m}_v+h(\mathbf{o}-0.5),\\
  \mathbf{f}^{dc} &= (\mathbf{c}-0.5)/C_0,\\
  \alpha &= \log\!\left(\bar p/(1-\bar p)\right),\\
  \mathbf{s} &= s_{\min}+(s_{\max}-s_{\min})\mathbf{l},\\
  \mathbf{q} &= \mathrm{normalize}(2\mathbf{r}-1).
  \end{align*}
  \item Append $(\mathbf{x},\alpha,\mathbf{f}^{dc},\mathbf{s},\mathbf{q})$ to $\hat{\mathcal{G}}$ and return the concatenated set after all active voxels are processed.
\end{enumerate}
\end{tcolorbox}

\section{Details of Overlap-Aware Tiled Inference}
\label{sec:supp_sliding_window}

\paragraph{Overview}
We generate scenes beyond one training crop using overlap-aware tiled inference. The procedure builds a large scene tile-by-tile with overlapping boundaries. For each spatial tile, the aligned patch from the supplied large-footprint satellite-view image provides conditioning at the same ground sampling resolution.

\paragraph{Tile Traversal and Overlap}
Tiles are arranged on the horizontal plane and visited in a BFS-spiral order starting from the center.
Each tile spans $32^3$ cells in the coarse sparse-structure grid, corresponding to a $256^3$ fine GS-Voxel grid. Neighboring tiles overlap by eight coarse-grid cells, giving a coarse-grid stride of 24.

\paragraph{Overlap Handling}
For tiles beyond the first, overlap regions are generated via repaint-style inpainting. Stored overlap latents from previously generated neighbors are noised to the current sampling time and used as constraints, while non-overlap regions are sampled freely under the boundary context.

\paragraph{Seam Reduction}
We use a three-stage schedule consistent with the main pipeline: coarse sparse structure is propagated first, geometry then predicts the fine active support, and attribute latents are finally generated and decoded into Gaussian parameters on that support. Shared noise in overlaps and feathered accumulation are used to reduce visible seams.

\paragraph{Post-processing}
We fill small holes in the coarse sparse-structure grid by nearest-donor copying and enforce the required subdivisions before geometry generation.

\paragraph{Settings}
We use 50 Euler steps per flow-matching stage with classifier-free guidance strength 3.0. The feathering validity offset is set to two coarse-grid cells.

\clearpage
\onecolumn

\section{Prompt for filtering}
\label{sec:prompt_filtering}

\begin{tcolorbox}[
  colback=white,
  colframe=black!35,
  boxrule=0.6pt,
  arc=2pt,
  left=4mm,right=4mm,top=3mm,bottom=3mm,
  width=\textwidth,
  breakable
]
\small
\begin{Verbatim}[
  fontsize=\small,
  breaklines=true,
  breakanywhere=true,
  breaksymbolleft={}
]
# Role
You are an extremely strict visual quality inspector for 3D Gaussian Splatting renders.
Be harsh on blurriness, but distinguish between "Scene Boundaries" and "Reconstruction Failures".

# Key Definitions
- Artifacts (fatal): blurry/smoky blobs, ghosting floaters, or "torn paper" holes that appear inside or over objects.
- Clean boundaries (acceptable): pure black regions at the edges or corners where the 3D model ends.
  If the transition from the scene to the black area is sharp and clean, it is NOT a failure.

# Guiding Principle
- Default for ~100% complete but soft images: ~0.60.
- Top scores (>0.85): ONLY for crystal clear, sharp textures.
- The "Boundary" Rule: A sharp, clean black corner/edge should only receive a minor penalty (e.g., -0.1),
  NOT a fatal ~0.20 score.

# Visual Checklist & Scoring (use the first matching description)
--- [CRITICAL FAILURE: Bad Reconstruction] ---
- Score 0.00 IF: a giant, blurry blob or "smoky smoke" obscures more than 30% of the important scene content.
- Score 0.25 IF: there are Internal Fatal Flaws:
  A) A "torn paper" hole inside a building or road.
  B) Messy, blurry "floaters" or smoky artifacts that overlap with main structures.

--- [PASS: Sharp Content, Possible Clean Boundaries] ---
- Score 0.55 IF: the image is complete (or has clean edges), but appears generally soft, hazy, or has geometric distortions.
  Details (windows, lines) are blurry.
- Score 0.70 IF: the image is structurally sound and sharp, but has a significant clean black boundary at the edges.
  The transition to black must be sharp.
- Score 0.80 IF: the image is very sharp and clear, with only a tiny, clean black sliver at a corner.
- Score 0.90+ IF: crystal clear across the entire view with 100% scene coverage and NO flaws.

# Output Instruction
Return ONLY valid JSON with this exact schema:
{ "score": <number in [0, 1]> }

Rules:
- Return exactly one score.
- score must be a number in [0, 1].
- No extra text, no markdown.
\end{Verbatim}
\end{tcolorbox}

\clearpage
\twocolumn

%% file: body/5_1_imagesupp.tex
\begin{figure*}[t]
  \centering
  \includegraphics[width=\linewidth]{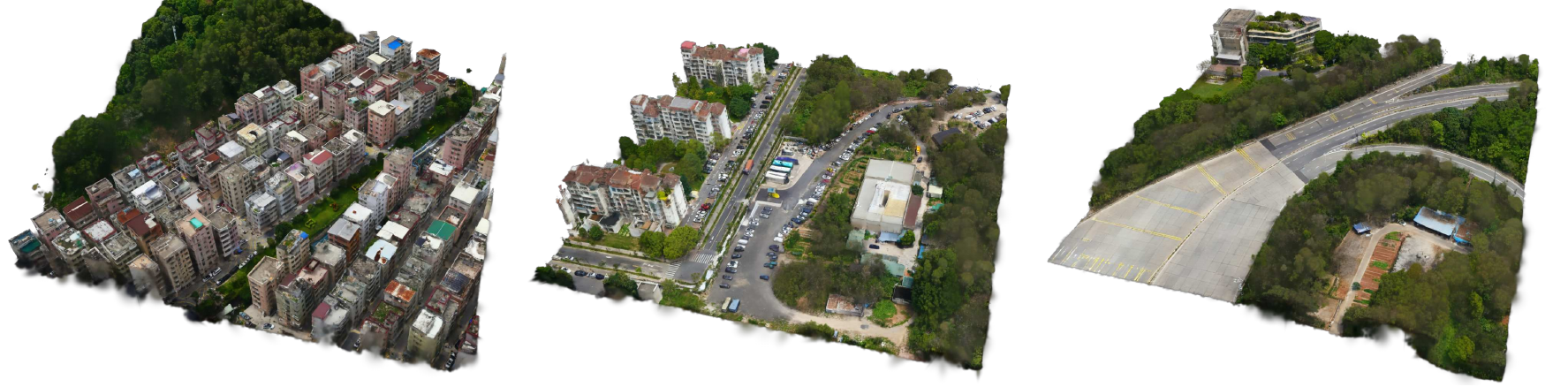}
  \Description{}
    \caption{Tile-level 3DGS generations conditioned on rendered satellite-view images.}
  \label{fig:tile_result}
\end{figure*}

\begin{figure*}[t]
  \centering
  \includegraphics[width=\textwidth]{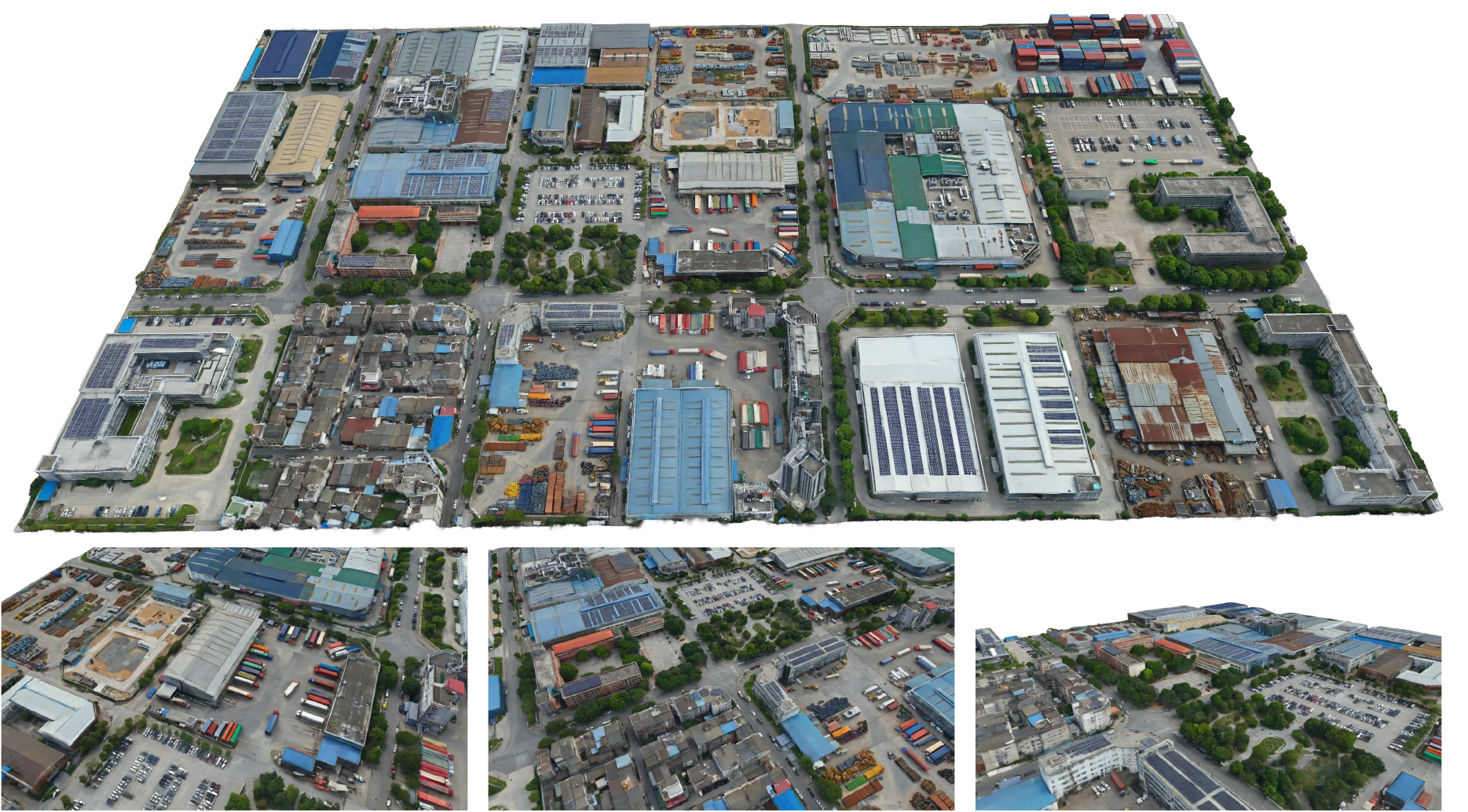}
  \Description{}
    \caption{Large-area 3DGS generation (1/2) from constructed satellite-view conditions.}
\vspace{1em}
\label{fig:large_scale_result1} 
\end{figure*}

\begin{figure*}[t]
  \centering
  \includegraphics[width=\textwidth]{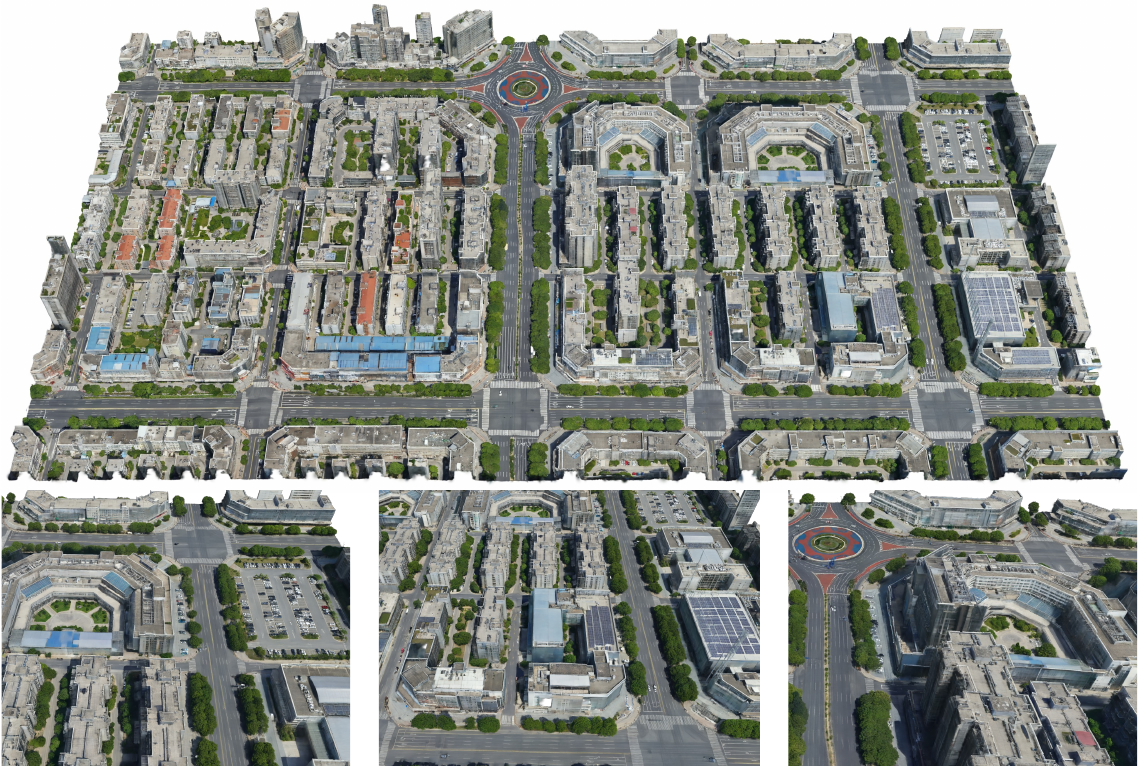}
  \Description{}
    \caption{Large-area 3DGS generation (2/2). Overlap-aware tiled inference produces a large-area 3DGS primitive set from the satellite-view condition.}
\vspace{1em}
\label{fig:large_scale_result2} 
\end{figure*}